\documentclass{midl} 

\usepackage{mwe} 
\usepackage[utf8]{inputenc} 
\usepackage[T1]{fontenc}    
\usepackage{hyperref}       
\usepackage{url}            
\usepackage{booktabs}       
\usepackage{amsfonts}       
\usepackage{nicefrac}       
\usepackage{microtype}      
\usepackage{tikz}
\usepackage{times}
\usepackage{graphicx}
\usepackage{amsmath}
\usepackage{amssymb}
\usepackage{multirow}
\usepackage{float}
\usepackage{stmaryrd}
\usepackage{xspace}
\usepackage{wrapfig}
\usepackage{caption}
\usepackage{booktabs}

\graphicspath{ {images/} }

\DeclareMathOperator{\mse}{MSE}
\DeclareMathOperator{\ncc}{NCC}
\DeclareMathOperator{\nccsup}{NCC_{sup}}
\DeclareMathOperator{\deepsim}{DeepSim}
\DeclareMathOperator{\deepsimseg}{DeepSim_{seg}}
\DeclareMathOperator{\deepsimae}{DeepSim_{ae}}

\DeclareMathOperator{\vgg}{VGG}
\newcommand{\I}{\ensuremath{\mathbf{I}}}
\newcommand{\J}{\ensuremath{\mathbf{J}}}
\newcommand{\A}{\ensuremath{\mathbf{A}}}
\newcommand{\B}{\ensuremath{\mathbf{B}}}
\newcommand{\Ss}{\ensuremath{\mathbf{S}}}

\newcommand{\pcoord}{\ensuremath{\mathbf{p}}}
\definecolor{grey}{gray}{0.55}

\newcommand{\transform}{\ensuremath{\Phi}}

\newcommand{\displacefield}{\ensuremath{\mathbf{u}}}

\newcommand{\domain}{\ensuremath{\Omega}}
\newcommand{\coord}{\ensuremath{\mathbf{x}}}

\newcommand{\dist}{\ensuremath{D}}
\newcommand{\reg}{\ensuremath{R}}
\newcommand{\loss}{\ensuremath{L}}

\jmlryear{2021}
\jmlrworkshop{Full Paper -- MIDL 2021}

\title[Semantic similarity metrics for learned image registration]{Semantic similarity metrics for learned image registration}

\midlauthor{\Name{Steffen Czolbe} \Email{per.sc@di.ku.dk}\\
\Name{Oswin Krause} \Email{oswin.krause@di.ku.dk}\\
\addr Department of Computer Science, University of Copenhagen, Denmark \AND
\Name{Aasa Feragen} \Email{afhar@dtu.dk}\\
\addr DTU Compute, Technical University of Denmark, Denmark}

\begin{document}

\maketitle

\begin{abstract}
We propose a semantic similarity metric for image registration. Existing metrics like Euclidean Distance or Normalized Cross-Correlation focus on aligning intensity values, giving difficulties with low intensity contrast or noise. Our approach learns dataset-specific features that drive the optimization of a learning-based registration model. We train both an unsupervised approach using an auto-encoder, and a semi-supervised approach using supplemental segmentation data to extract semantic features for image registration. Comparing to existing methods across multiple image modalities and applications, we achieve consistently high registration accuracy. A learned invariance to noise gives smoother transformations on low-quality images. Code and experiments are available at \href{https://github.com/SteffenCzolbe/DeepSimRegistration}{\texttt{github.com/SteffenCzolbe/DeepSimRegistration}}.
\end{abstract}


\section{Introduction}

Deformable registration, or nonlinear image alignment, is a fundamental tool in medical imaging. Registration models find correspondences between a set of images and derive a geometric transformation to align them. Most algorithmic and deep learning-based methods solve the registration problem by the minimization of a loss function, consisting of a similarity metric and a regularization term ensuring smoothness of the transformation. The similarity metric is essential to the optimization; it judges the quality of the match between registered images and has a strong influence on the result. 

Pixel-based similarity metrics like euclidean distance and patch-wise cross-correlation are well explored within algorithmic~\cite{Avants2011, Avants2008, faisalBeg2005, Rueckert1999, thirion1998, Vercauteren2007} and deep learning based~\cite{Alven2019, balakrishnan2019voxel, dalca2018unsuperiveddiff, Dalca2019, DeVos2019, Liu2019, Lee2019a, Hu2019, Hu2019a, Yang2017, Xu2019} image registration.
These metrics assume that if the image intensities are aligned, or strongly correlated, the images are well aligned. Each choice of metric adds additional assumptions on the characteristics of the specific dataset. Thus, a common methodological approach is to trial registration models with multiple different pixel-based metrics, and choose the metric performing best on the dataset \cite{balakrishnan2019voxel, Hu2019}.

The shortcomings of pixel-based similarity metrics, such as blur in the generated images, have been studied substantially in the image generation community \cite{hou2017deep, Zhang2018} and have been superseded by deep similarity metrics approximating human visual perception. Here, features are extracted from neural networks pre-trained on image-classification tasks~\cite{deng2009imagenet}. Performance can further be improved by fine-tuning the features to human perception \cite{Czolbe2020, Zhang2018}, leading to generative models that produce photo-realistic images.
We propose to apply deep similarity metrics within image registration to achieve a similar increase of performance for registration models.

\paragraph{Contribution} We propose a data-driven similarity metric for image registration based on the alignment of semantic features. We explore both unsupervised (using auto-encoders) and semi-supervised methods (using a segmentation model) to learn filters of semantic importance to the dataset. We use the learned features to construct a similarity metric used for training a registration model, and validate our approach on three biomedical datasets of different image modalities and applications. Across all datasets, our method achieves consistently high registration accuracy, outperforming even metrics utilizing supervised information on two out of three datasets, while yielding an inconclusive result on the third one. Our models learn to ignore noisy image patches, leading to smoother transformations on low-quality data.

\section{Background \& related work} \label{sec:background}

\paragraph{Image registration}

Most image registration frameworks model the problem as finding a transformation $\transform \colon \domain \to \domain$ that aligns a moving image $\I \colon \domain \to \mathbb{R}$ to a fixed image $\J \colon \domain \to \mathbb{R}$. The morphed source image, obtained by applying the transformation, is expressed by function composition as $\I \circ \transform$. The domain $\domain$ denotes the set of all coordinates $\coord \in \mathbb{R}^d$ within the image\footnote{While the domain $\domain$ is continuous in $\mathbb{R}^d$, recorded images and computations thereon are discrete. For simplicity of notation, we denote both the continuous and discrete domain as $\domain$. We implement $\sum_{\pcoord \in \domain}$ as a vectorized operation over the discrete pixel/voxel-coordinates and calculate $|\domain|$ as the total count of discrete pixels/voxels of the image. The transformation $\transform$ is implemented as a map from a discrete domain to a continuous one, and the sampling of continuous points from a discrete image is implemented via bi-/tri-linear interpolation.}. Images record intensity at discrete pixel-coordinates $\pcoord$ but can be viewed as a continuous function by interpolation.

The transformation is found through iterative algorithms \cite{Avants2008, faisalBeg2005, Rueckert1999, thirion1998, Vercauteren2007}, or predicted with learning based techniques \cite{balakrishnan2019voxel, dalca2018unsuperiveddiff, DeVos2019, Liu2019, Lee2019a, Hu2019a, Yang2017, Xu2019}. In both cases, the optimal transformation is found by minimization of a similarity measure $\dist$ and a $\lambda$-weighted regularizer $\reg$, expressed via the loss function
\begin{equation} \label{eq:reg-loss}
\loss(\I, \J, \transform) = \dist(\I \circ \transform, \J) + \lambda \reg(\transform) \enspace .
\end{equation}
As many non-linear transformation models are over-parametrized and have multiple minima, regularization is necessary. Smooth transformation fields, that avoid folds or gaps, are assumed to be physically plausible and encouraged by the regularizer. We use the \textit{diffusion regularizer} throughout this paper, which is defined as
\begin{equation}  \label{eq:diffusion-regularizer}
\transform(\pcoord) = \pcoord + \displacefield(\pcoord) , \enspace\reg(\transform) = \sum_{\pcoord \in \domain} \Vert \nabla \displacefield(\pcoord) \Vert^2 \enspace ,
\end{equation}
with the gradient of the displacement field $\nabla \displacefield(\pcoord)$ approximated via finite differences. 

\paragraph{Similarity metrics for image registration}
Denote by $\dist$ the (dis-)similarity between the morphed moving image $\I \circ \transform$ and the fixed image $\J$. Pixel-based  metrics are well explored within algorithmic image registration, a comparative evaluation is given by \citet{Avants2011}. We briefly recall the two most popular choices, \textit{mean squared error} ($\mse$) and \textit{normalized cross correlation} ($\ncc$).
The pixel-wise MSE is intuitive and computationally efficient. It is derived from maximizing the negative log-likelihood of a Gaussian normal distribution, making it an appropriate choice under the assumption of Gaussian noise. On a grid of discrete points $\pcoord$ from domain $\domain$, the $\mse$ is defined as $\mse(\I \circ \transform, \J) = \frac{1}{|\domain|} \sum_{\pcoord \in \domain}  \Vert \I \circ \transform(\pcoord) - \J(\pcoord) \Vert^2$.

Patch-wise normalized cross correlation is robust to variations in brightness and contrast, making it a popular choice for images recorded with different acquisition tools and protocols, or even across image modalities. For two image patches $\A, \B$, represented as column-vectors of length $N$ with patch-wise means $\bar{\A}, \bar{\B}$ and variance $\sigma^2_{\A}, \sigma^2_{\B}$, it is defined as
\begin{equation} \label{eq:ncc}
\ncc_{\text{patch}}(\A, \B) = \sum_{n=1}^{N} \frac{(\A_n - \bar{\A}) (\B_n - \bar{\B})}{\sigma_\A \sigma_\B} \enspace .
\end{equation}
The Patch-wise similarities are then averaged over the image \cite{gee1993elastically, Avants2008}. 
Note that an alternative, computationally efficient variant of $\ncc$  is sometimes used in image registration \cite{Avants2011}. If annotations are available, these unsupervised similarity measures can be extended by a \textit{supervised} component to measure both intensity differences and the alignment of annotated label maps \cite{balakrishnan2019voxel}.

\paragraph{Learned similarity metrics for image registration}

While deep-learning-based image registration has received much interest recently, similarity metrics utilizing the compositional and data-driven advantages of neural networks remain under-explored. Some current works explore how to incorporate scale-space into learned registration models, but similarity metrics remain pixel-based \cite{Hu2019, Li2018}. Learned similarity metrics are proposed by \citet{Haskins2019a} and \citet{Krebs2017}, but both approaches require ground truth registration maps, which are either synthetically generated or manually created by a medical expert. \citet{Lee2019a} propose to learn annotated structures of interest as part of the registration model to aid alignment, but the method discards sub-regional and non-annotated structures.

Learned common data representations and similarity metrics are frequently used in multi-modal image registration \cite{Heinrich2012,Chen2016,Simonovsky2016,Pielawski2020a}. While these approaches learn common representations from well-aligned images of multiple modalities, we aim to find a semantically augmented representation of images of a single modality.

\section{Method}

\paragraph{A discussion of NCC}
Our design of a semantic similarity metric starts by examining the popular $\ncc$ metric. We see that $\ncc$ between image patches $\A$ and $\B$ is equivalent to the cosine-similarity between the corresponding mean-centered vectors $f(\A) = \A - \bar{\A}$ and $f(\B) = \B- \bar{\B}$:
\begin{equation}\label{eq:ncc_with_cossim}
\ncc_{\text{patch}}(\A, \B) =  \frac{ \big< f(\A), f(\B) \big>}{\Vert f(\A)\Vert \Vert f(\B) \Vert}\enspace,
\end{equation}
with scalar product $\langle \cdot , \cdot \rangle$ and euclidean norm $\Vert \cdot \Vert$ . Thus, an alternative interpretation of the $\ncc$ similarity measure is the cosine-similarity between two feature descriptors in a high-dimensional space. The descriptor is given by the intensity values of a centered image patch centered at a pixel~$\pcoord$. We will construct a similar metric, using semantic feature descriptors instead.

\paragraph{A semantic similarity metric for image registration}
To align areas of similar semantic value, we propose a similarity metric based on the agreement of semantic feature representations of two images. Semantic feature maps are obtained by a \textit{feature extractor}, which is pre-trained on a surrogate task. To capture alignment of both localized, concrete features, and global, abstract ones, we calculate the similarity at multiple layers of abstraction. Given a set of feature-extracting functions $F^l \colon \mathbb{R}^{\domain \times C} \to \mathbb{R}^{\domain_l \times C_l}$ for $L$ layers, we define
\begin{align}
\deepsim(\I \circ \transform, \J) = \frac{1}{L} \sum_{l=1}^{L} \frac{1}{| \domain_l |} \sum_{{\pcoord} \in \domain_l}
\frac{\big< F^l_{\pcoord}(\I \circ \transform), F^l_{\pcoord}(\J)\big>}{\Vert  F^l_{\pcoord}(\I \circ \transform) \Vert \Vert  F^l_{\pcoord}(\J) \Vert} \enspace ,
\end{align}
where $F^l_{\pcoord}(\J)$ denotes the $l^{th}$ layer feature extractor applied to image $\J$, at spatial coordinate $\pcoord$. It is a vector of $C_l$ output channels, and the spatial size of the $l^{th}$ feature map is denoted by $| \domain_l |$. 

Similarly to $\ncc$, the neighborhood of the pixel is considered in the metric, as we compose $F^l$ of convolutional filters with increasing receptive area of the composition. In contrast, it is not necessary to zero-mean the feature descriptors, as the semantic feature representations are trained to be robust to variances in image brightness present in the training data.

\begin{figure}
	\centering
	\def\svgwidth{0.9 \linewidth}
	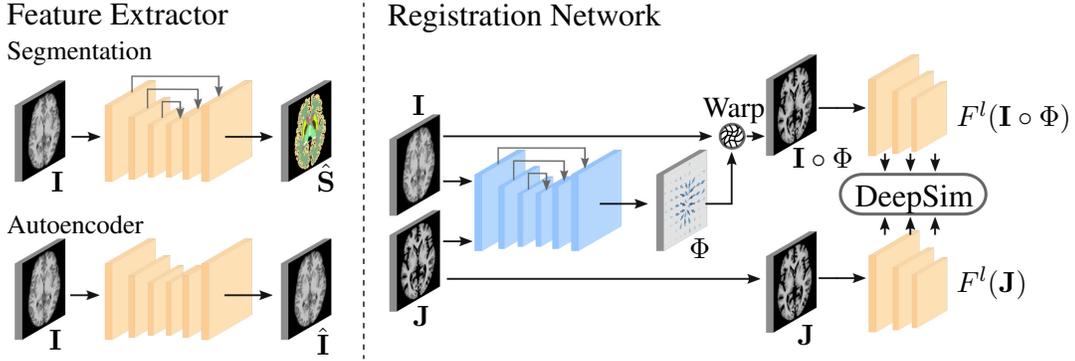
	\caption{Two-step training: First, the feature extractor (yellow) is trained. We test both a segmentation model, and an auto-encoder. Next, the feature extractor weights are frozen and used in the loss computation of the registration network (blue).}
	\label{fig:model}
\end{figure}

\paragraph{Feature extraction}
To aid registration, the functions $F^l(\cdot)$ should extract features of semantic relevance for the registration task, while ignoring noise and artifacts. We extract features from the encoding branch of networks trained on two surrogate tasks.

First, if segmentation masks are available, we can learn features on a supplementary segmentation task. Segmentation models excel at learning relevant kernels for the data while attaining invariance towards non-predictive features like noise, but require an annotated dataset for training.  We denote the proposed similarity metric with feature extractors conditioned on this task as $\deepsimseg$.

Second, we can learn an abstract feature representation of the dataset in an unsupervised setting with auto-encoders. Auto-encoders learn efficient data encoding by training the network to ignore signal noise. A benefit of this approach is that no additional annotations are required.  While variational methods for encoding tasks have several advantages, we choose a deterministic auto-encoder for its simplicity and lack of hyperparameters. We denote the similarity metric with feature extractors conditioned on this task as $\deepsimae$.

\begin{figure}
	\centering
	\subfigure[Qualitative comparison]
	{
		\label{fig:a}
		\begin{tikzpicture}
		\node[anchor=south west,inner sep=0] (image) at (0,0) {%
			\includegraphics[width=0.832\textwidth]{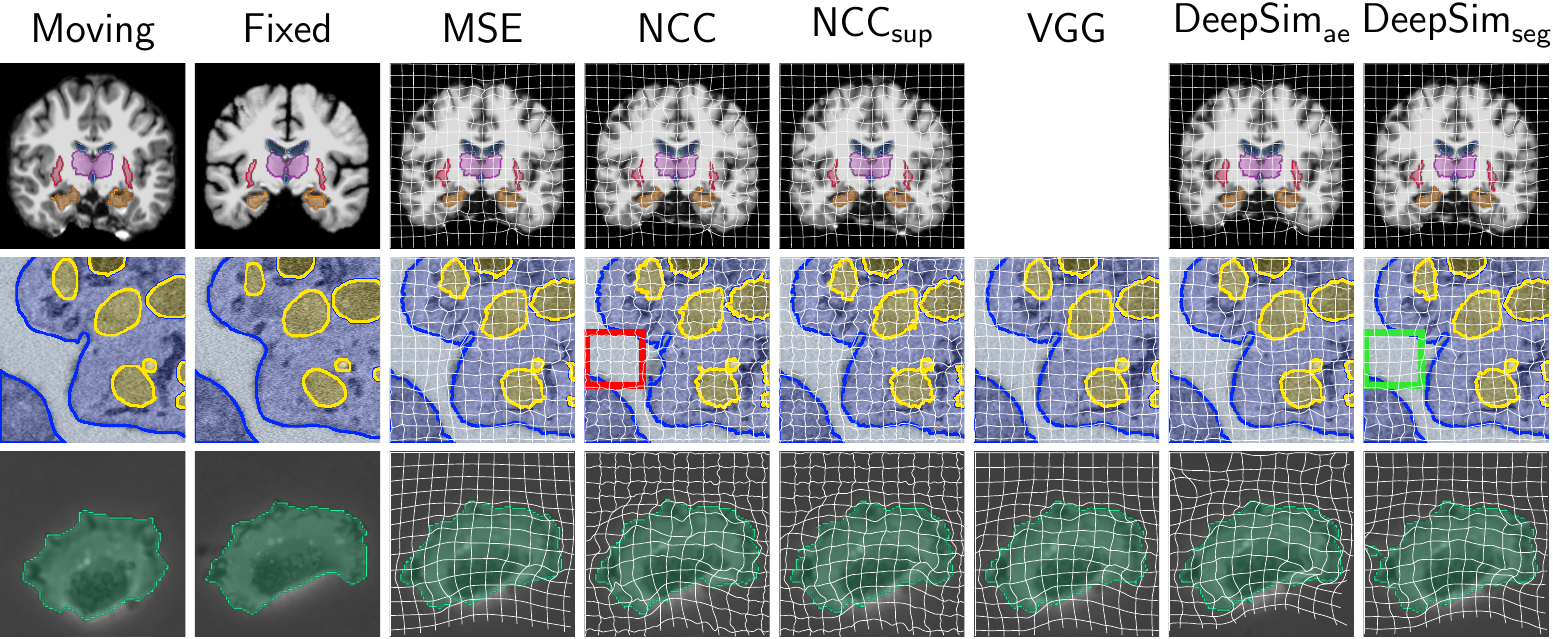}
		};
		\draw (7.97,4.71) -- (9.46,4.71) -- (9.46,3.2) -- (7.97,3.2) -- (7.97,4.71);		\node[align=center, scale=0.75] at (8.715,3.955) { N/A for \\ 3D Data};
		\end{tikzpicture}
	}
	\subfigure[Detail view]
	{
		\label{fig:samples_detail}
		\includegraphics[width=.135\textwidth]{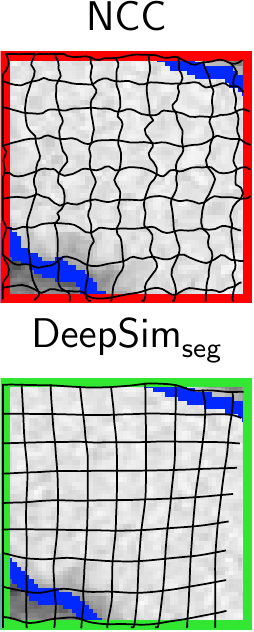}
	}
	\caption{Left: Qualitative comparison of registration models. Rows: Datasets Brain-MRI, Platelet-EM, PhC-U373. Columns 3-8: Registration models trained with various similarity metrics. Right: Detail view of highlighted noisy background areas on the Platelet-EM dataset. Select segmentation classes annotated in color. The transformation is visualized by morphed grid-lines. }
	\label{fig:samples}
\end{figure} 

\newpage
\section{Experiments}

\begin{wrapfigure}{r}{0.5\textwidth}
\includegraphics[width=1\linewidth]{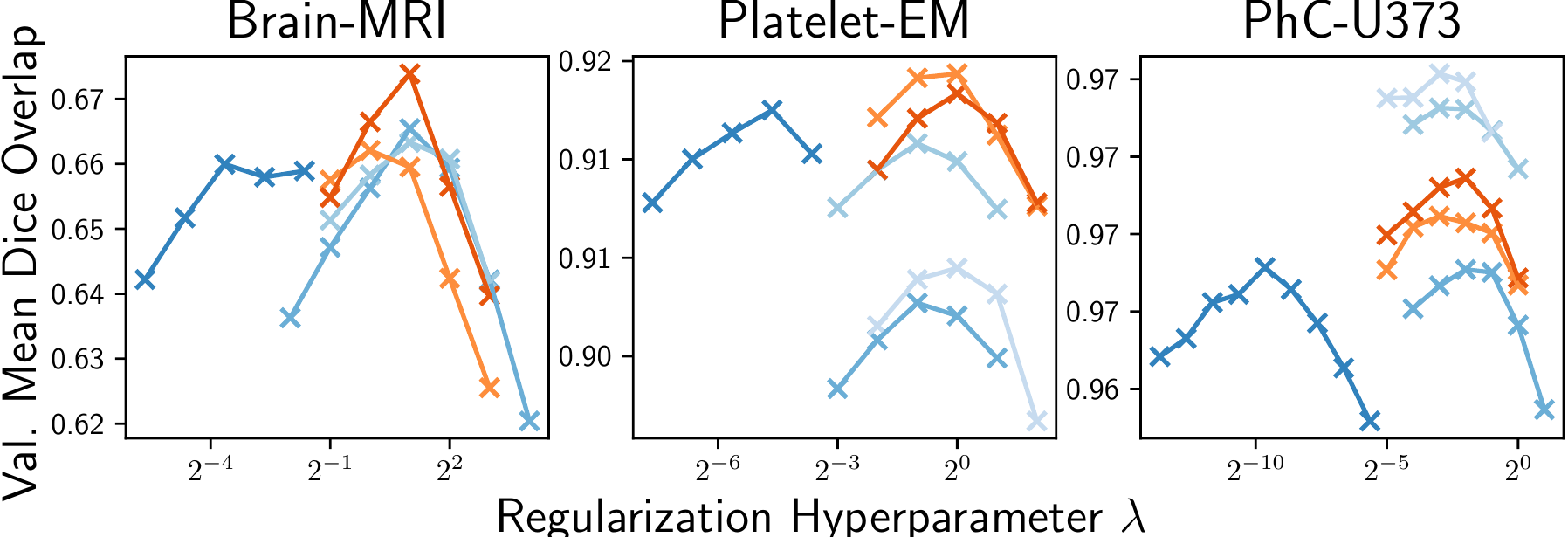}
\caption{Hyperparameter tuning. For each model, we select the regularizer weighting factor $\lambda$ with the highest validation mean dice overlap for further evaluation. Color scheme follows Figure~\ref{fig:testscore}.}
\label{fig:hparams}
\end{wrapfigure}

We empirically compare registration models trained with the unsupervised $\deepsimae$ and semi-supervised $\deepsimseg$ to the baselines $\mse$, $\ncc$, $\nccsup$ ($\ncc$ with supervised information), and $\vgg$ (a VGG-net based deep similarity metric from image generation). Our implementation of the baseline methods follows \citet{Avants2011}, \citet{balakrishnan2019voxel}, and \citet{hou2017deep}.

As our goal is not to advance the state of the art for any particular registration task, but instead to explore the value of our loss-function in a generic setting, we use well-established 2D and 3D U-Net \cite{Ronneberger2015} architectures for both registration and segmentation models, sketched in Figure~\ref{fig:model}. For the auto-encoder task, we use the same architecture, but without the shortcut connections. Each network consists of three encoder and decoder stages. Each stage consists of one batch normalization \cite{Ioffe2015}, two convolutional, and one dropout layer \cite{Gal2016}. After the final decoder step, we smooth the model output with three more convolutional layers. We experimented with deeper architectures but found they do not increase performance. The activation function is LeakyReLu throughout the network, Softmax for the final layer of the segmentation network, Sigmoid for the final layer of the auto-encoder, and linear for the final layer of the registration network. The stages have $64, 128, 256$ channels for 2d datasets, and $32, 64, 128$ channels for 3d. 

The segmentation model is trained with a cross-entropy loss function, the auto-encoder with the mean squared error, and the registration network with the loss given by Eq.~\ref{eq:reg-loss}. The optimization algorithm for all models is ADAM \cite{Kingma2015}, the initial learning rate is $10^{-4}$, decreasing by a factor of $10$ each time the validation loss plateaus. All models are trained until convergence. Due to the large 3D volumes involved, the choice of batch-size is often limited by available memory. We sum gradients over multiple passes to arrive at effective batch-sizes of 3-5 samples. The hyperparameter $\lambda$ is tuned for each model independently on the validation set, we plot the validation score of tested values in Figure~\ref{fig:hparams}.

To show that our approach applies to a variety of registration tasks, we validate it on three 2D and 3D datasets of different modalities: 4000 T1-weighted \textit{Brain-MRI} scans \cite{di2014autism, lamontagne2019oasis}, 74 slices of human blood cells of the \textit{Platelet-EM} dataset \cite{Quay2018}, and 230 time-steps of the cell-tracking video \textit{PhC-U373} \cite{Ulman2014, Ulman2017}. All datasets are pre-aligned through affine transformations and split into train, validation, and test sections. We augment each pair of images with random affine transformations during training.

\begin{figure}
\centering
\includegraphics[width=0.9\linewidth]{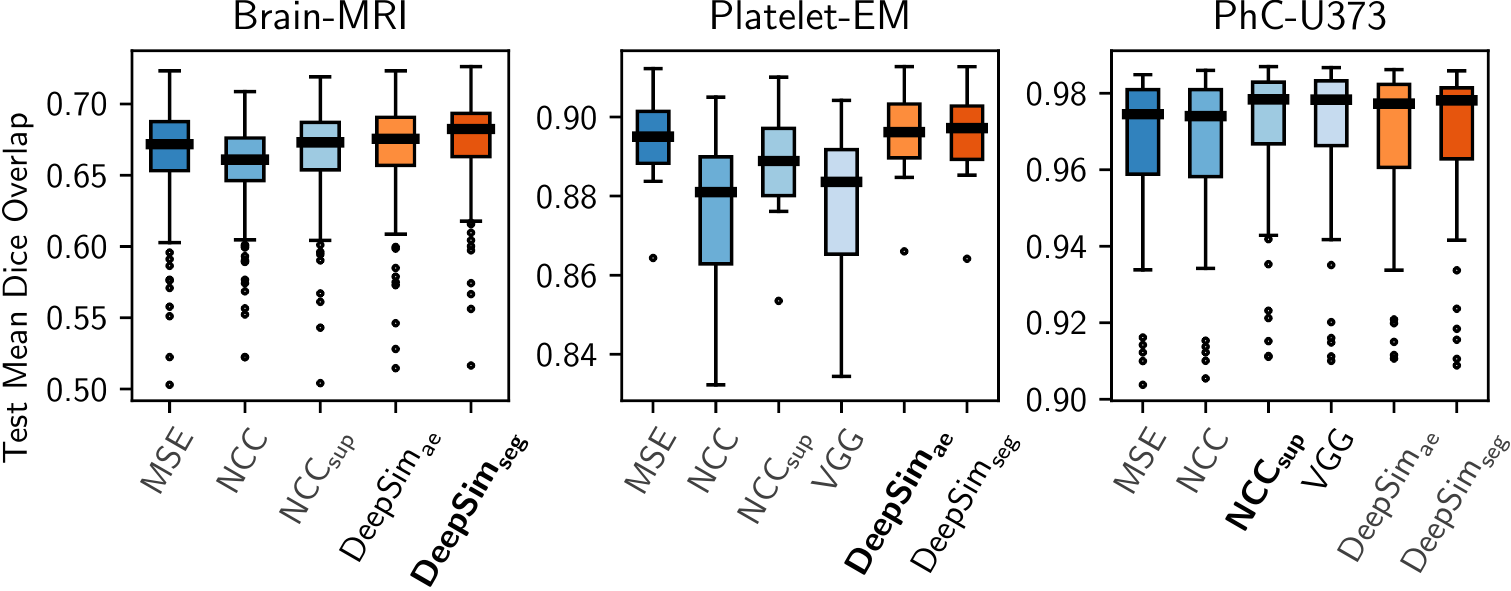}
\caption{Mean dice overlap of registration models on the test set. Baselines in shades of blue, ours in red. Label of the best metric in bold, 2nd best black, others in grey. Boxplot with median, quartiles, deciles and outliers.}
\label{fig:testscore}
\end{figure}
\begin{table}[]
	\centering
	\begingroup 
	\setlength{\tabcolsep}{5pt} 
	\begin{tabular}{@{}llrrrrrrrr@{}}
		\toprule[0.1em]
		Dataset                      & \multicolumn{1}{c}{Method} & \multicolumn{8}{c}{Baseline}                                                                                                                                                                                 \\ \cmidrule(r){1-1} \cmidrule(lr){2-2} \cmidrule(l){3-10}
		&                              & \multicolumn{2}{c}{$\mse$}                        & \multicolumn{2}{c}{$\ncc$}                        & \multicolumn{2}{c}{$\ncc_{\text{sup}}$}           & \multicolumn{2}{c}{$\vgg$}                        \\ \cmidrule(lr){3-4} \cmidrule(lr){5-6} \cmidrule(lr){7-8} \cmidrule(l){9-10}
		&                              & \multicolumn{1}{c}{$p$} & \multicolumn{1}{c}{$d$} & \multicolumn{1}{c}{$p$} & \multicolumn{1}{c}{$d$} & \multicolumn{1}{c}{$p$} & \multicolumn{1}{c}{$d$} & \multicolumn{1}{c}{$p$} & \multicolumn{1}{c}{$d$} \\ \addlinespace 
		\multirow{2}{*}{Brain-MRI}                    & $\deepsimae$       & $<\!0.001$                 & $0.14$                  & $<\!0.001$                 & $0.43$                  & $<\!0.001$                 & $0.10$                  & \multicolumn{1}{c}{--}                & \multicolumn{1}{c}{--}                   \\
		& $\deepsimseg$      & $<\!0.001$                 & $0.30$                  & $<\!0.001$                 & $0.60$                  & $<\!0.001$                 & $0.25$                  & \multicolumn{1}{c}{--}                & \multicolumn{1}{c}{--}                \\ \addlinespace 
		\multirow{2}{*}{Platelet-EM} & $\deepsimae$       & $\color{grey} 0.016$                 & $ 0.12$                  & $<\!0.001$                 & $1.14$                  & $<\!0.001$                 & $0.51$                  & $<\!0.001$                 & $1.06$                  \\
		& $\deepsimseg$      & $\color{grey} 0.034$                 & $0.10$                  & $<\!0.001$                 & $1.12$                  & $<\!0.001$                 & $0.49$                  & $<\!0.001$                 & $1.04$                  \\ \addlinespace 
		\multirow{2}{*}{PhC-U373}    & $\deepsimae$       & $<\!0.001$                 & $0.10$                  & $<\!0.001$                 & $0.11$                  & $<\!0.001$                 & $\color{grey} -0.06$                  & $<\!0.001$                 & $\color{grey} -0.05$                  \\
		& $\deepsimseg$      & $<\!0.001$                 & $0.12$                  & $<\!0.001$                 & $0.13$                  & $0.002$                 & $\color{grey} -0.03$                  & $\color{grey} 0.003$                 & $\color{grey} -0.02$                  \\ \bottomrule[0.1em]
	\end{tabular}
	\endgroup
	\caption{Results of the statistical significance test, performed with the Wilcoxon signed rank test for paired samples. Effect size measured with Cohen's d. Statistically insignificant results (significance level 0.05, Bonferroni-corrected to $p>0.002$) and very small effect sizes ($|d| <0.1$) in grey.}
	\label{tab:statistical-tests}
\end{table}

\section{Results} 
\paragraph{Registration accuracy}
We measure the mean S{\o}rensen Dice coefficient on the unseen test-set in Figure~\ref{fig:testscore}. Statistical significance testing of the results is performed with the Wilcoxon signed rank test for paired samples. A significance level of $5\%$ gives Bonferroni-adjusted significance threshold $p=0.002$. We further measure the effect size with Cohen's d and show the results in Table \ref{tab:statistical-tests}.
Models trained with our proposed $\deepsimae$ and $\deepsimseg$ outperform all baselines on the Brain-MRI and Platelet-EM datasets, with strong statistical significance and effect size. On the \mbox{PhC-U373} dataset, all models achieve high dice-overlaps of $>0.97$. 

We monitor the mean dice overlap during training. The training accuracy is, with few exceptions, similar to the test accuracy, indicating that results generalize well. The empirical convergence speeds of the tested metrics differ. We observe that the $\deepsimseg$ converges faster than the baselines, especially in the first few epochs of training. See appendix \ref{appendix:convergence} for a convergence plot.

\paragraph{Qualitative examples \& transformation grids}
We plot the fixed and moving images $\I, \J$ and the morphed image $\I \circ \transform$ for each similarity metric model along with a more detailed view of a noisy patch of the Platelet-EM dataset in Figure~\ref{fig:samples}, and perform a quantitative analysis of the transformation in Appendix~\ref{appendix:transformation}. On models trained with the baselines, we find strongly distorted transformation fields in noisy areas of the images. In particular, models trained with $\ncc$ and $\ncc_{\text{sup}}$ produce very irregular transformations, despite careful tuning of the regularization-hyper-parameter. The model trained by $\deepsimseg$ is more invariant towards noise.

\begin{figure}
	\centering
	\includegraphics[width=1\linewidth]{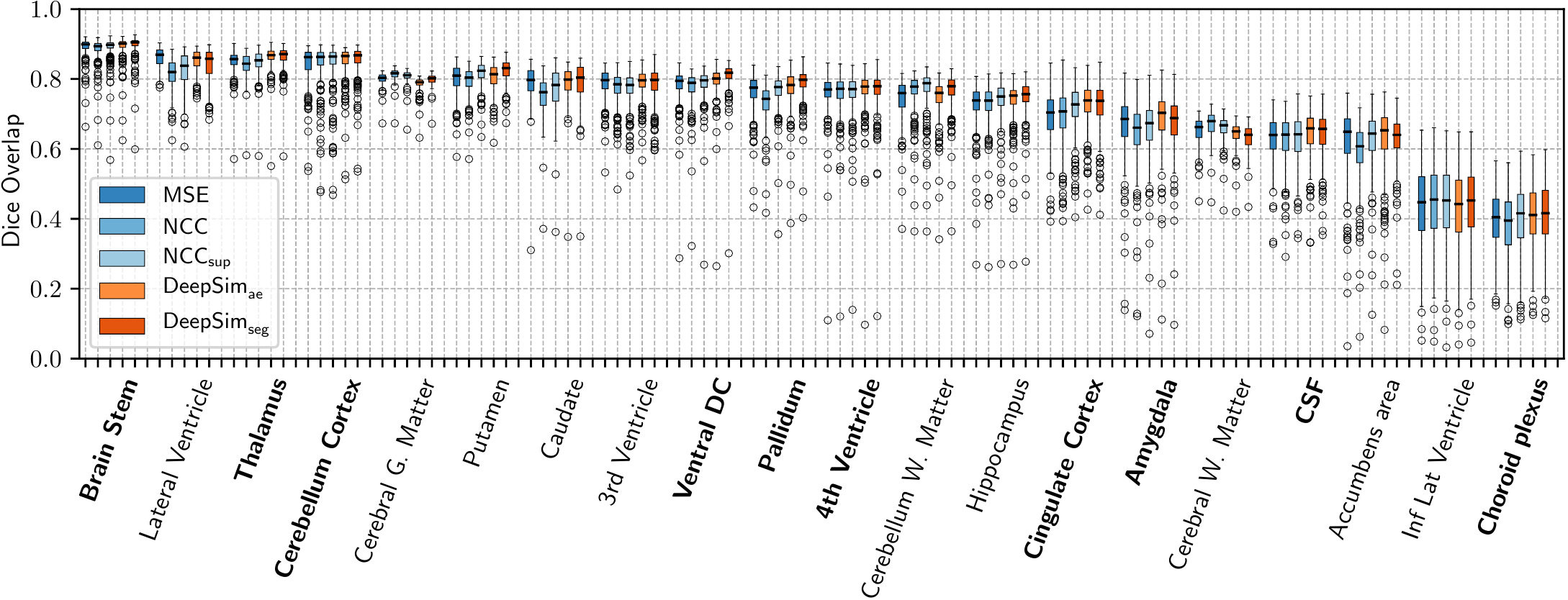}
	\caption{Dice overlaps of the anatomical regions of the Brain-MRI dataset. Baselines in shades of blue, our methods in red. Bold labels for regions where \textit{both} of our methods score higher than all baselines. We combined labels of the left and right brain hemispheres into a single class. The boxplot shows median, quartiles, deciles and outliers.}
	\label{fig:test_score_per_class_brain}
\end{figure}

\paragraph{Anatomical regions}
The Brain-MRI dataset is annotated with the anatomical regions of the brain. We plot the dice overlap per region in a boxplot in Figure~\ref{fig:test_score_per_class_brain}, and highlight regions where both of our metrics perform better than all baselines bold. Baseline methods (blue) perform very similar, despite $\ncc_{\text{sup}}$ as a supervised metric requiring more information over the unsupervised $\mse$ and $\ncc$.

\section{Discussion \& Conclusion}
Registration models trained with $\deepsim$ achieve high registration accuracy across datasets, allowing improved downstream analysis and diagnosis. 
Its consistency makes testing multiple traditional metrics unnecessary; instead of empirically determining whether $\mse$ or $\ncc$ captures the characteristics of a data-set best, we can use $\deepsim$ to learn the relevant features from the data. 

Our experiments show that the availability of annotated anatomical regions can help in learning semantic features, but is not a necessity. Gains in performance in deep learning often require large, annotated datasets, which are expensive and time-consuming to obtain in biomedical settings. The feature extractor of $\deepsimae$ was trained on the unsupervised autoencoding task, requiring no additional data aside from the intensity images to be registered. It performed similarly to the semi-supervised $\deepsimseg$ on two of the three datasets and outperformed the baselines.

The analysis of noisy patches in Figure~\ref{fig:samples} and Appendix~\ref{appendix:transformation} highlights a learned invariance to noise. The pixel-based similarity metrics are distracted by artifacts, leading to overly-detailed transformation fields. Models trained with $\deepsim$ do not show this problem. While smoother transformation fields can be obtained for all metrics by strengthening the regularizer, this would negatively impact the registration accuracy of anatomically significant regions. Accurate registration of noisy, low-quality images allows for shorter acquisition time and reduced radiation dose in medical applications.

A weakness of $\deepsim$ is the need to train a separate model for feature extraction. The design, training, and testing of a second model takes additional resources, and the presented approach necessitates a dataset to train the feature extractor with. In the context of deep learning-based registration, $\deepsimae$ requires no additional data to what is required to train the registration model, while $\deepsimseg$ requires additional label maps. Compared to algorithmic registration methods, which optimize the registration map for each pair of input images separately, any deep-learning approach requires additional data.

$\deepsim$ is a general metric, applicable to image registration tasks of all modalities and anatomies. Beyond the presented datasets, our good results in the presence of noise let us hope that $\deepsim$ will improve registration accuracy in domains such as low dose CT, ultrasound, or microscopy, where details are often hard to identify, and image quality is poor. We further emphasize that the application of $\deepsim$ is limited neither to deep learning nor to image registration. In Algorithmic image registration, a similarity-based loss is minimized via gradient descent-based methods. $\deepsim$ can be applied to drive algorithmic methods, improving their performance by aligning deep, semantic feature embeddings. Similarly, $\deepsim$ could be used for other image regression tasks, such as image synthesis, -translation, or -reconstruction.

\newpage
\midlacknowledgments{This work was funded in part by the Novo Nordisk Foundation through the Center for Basic Machine Learning Research in Life Science (grant no. 0062606), and in part through the Lundbeck Foundation (grant no. R218-2016-883). We further thank Matthew Quay, the Cell Tracking Challenge, and the Cancer Imaging Archive for the provision of the datasets.}

\bibliography{czolbe21}

\begin{thebibliography}{38}
\providecommand{\natexlab}[1]{#1}
\providecommand{\url}[1]{\texttt{#1}}
\expandafter\ifx\csname urlstyle\endcsname\relax
  \providecommand{\doi}[1]{doi: #1}\else
  \providecommand{\doi}{doi: \begingroup \urlstyle{rm}\Url}\fi

\bibitem[Alv{\'{e}}n et~al.(2019)Alv{\'{e}}n, Heurling, Smith, Strandberg,
  Sch{\"{o}}ll, Hansson, and Kahl]{Alven2019}
Jennifer Alv{\'{e}}n, Kerstin Heurling, Ruben Smith, Olof Strandberg, Michael
  Sch{\"{o}}ll, Oskar Hansson, and Fredrik Kahl.
\newblock {A Deep Learning Approach to MR-less Spatial Normalization for Tau
  PET Images}.
\newblock In \emph{International Conference on Medical Image Computing and
  Computer-Assisted Intervention}, pages 355--363. 2019.

\bibitem[Avants et~al.(2008)Avants, Epstein, Grossman, and Gee]{Avants2008}
B.~B. Avants, C.~L. Epstein, M.~Grossman, and J.~C. Gee.
\newblock {Symmetric diffeomorphic image registration with cross-correlation:
  Evaluating automated labeling of elderly and neurodegenerative brain}.
\newblock \emph{Medical Image Analysis}, 12\penalty0 (1):\penalty0 26--41,
  2008.

\bibitem[Avants et~al.(2011)Avants, Tustison, Song, Cook, Klein, and
  Gee]{Avants2011}
Brian~B. Avants, Nicholas~J. Tustison, Gang Song, Philip~A. Cook, Arno Klein,
  and James~C. Gee.
\newblock {A reproducible evaluation of ANTs similarity metric performance in
  brain image registration}.
\newblock \emph{NeuroImage}, 54\penalty0 (3):\penalty0 2033--2044, 2011.

\bibitem[Balakrishnan et~al.(2019)Balakrishnan, Zhao, Sabuncu, Guttag, and
  Dalca]{balakrishnan2019voxel}
Guha Balakrishnan, Amy Zhao, Mert~R. Sabuncu, John Guttag, and Adrian~V. Dalca.
\newblock {VoxelMorph: A Learning Framework for Deformable Medical Image
  Registration}.
\newblock \emph{IEEE Transactions on Medical Imaging}, 38\penalty0
  (8):\penalty0 1788--1800, 2019.

\bibitem[Chen et~al.(2016)Chen, Duan, Houthooft, Schulman, Sutskever, and
  Abbeel]{Chen2016}
Xi~Chen, Yan Duan, Rein Houthooft, John Schulman, Ilya Sutskever, and Pieter
  Abbeel.
\newblock {InfoGAN: Interpretable Representation Learning by Information
  Maximizing Generative Adversarial Nets}.
\newblock In \emph{Advances in Neural Information Processing Systems}, 2016.

\bibitem[Czolbe et~al.(2020)Czolbe, Krause, Cox, and Igel]{Czolbe2020}
Steffen Czolbe, Oswin Krause, Ingemar Cox, and Christian Igel.
\newblock {A Loss Function for Generative Neural Networks Based on Watson's
  Perceptual Model}.
\newblock \emph{Advances in Neural Information Processing Systems}, 2020.

\bibitem[Dalca et~al.(2018)Dalca, Balakrishnan, Guttag, and
  Sabuncu]{dalca2018unsuperiveddiff}
Adrian~V. Dalca, Guha Balakrishnan, John Guttag, and Mert~R. Sabuncu.
\newblock {Unsupervised Learning for Fast Probabilistic Diffeomorphic
  Registration}.
\newblock \emph{Medical Image Computing and Computer Assisted Intervention},
  pages 729--738, 2018.

\bibitem[Dalca et~al.(2019)Dalca, Rakic, Guttag, and Sabuncu]{Dalca2019}
Adrian~V. Dalca, Marianne Rakic, John Guttag, and Mert~R. Sabuncu.
\newblock {Learning Conditional Deformable Templates with Convolutional
  Networks}.
\newblock \emph{Advances in Neural Information Processing Systems}, pages
  806----818, 2019.

\bibitem[de~Vos et~al.(2019)de~Vos, Berendsen, Viergever, Sokooti, Staring, and
  I{\v{s}}gum]{DeVos2019}
Bob~D. de~Vos, Floris~F. Berendsen, Max~A. Viergever, Hessam Sokooti, Marius
  Staring, and Ivana I{\v{s}}gum.
\newblock {A deep learning framework for unsupervised affine and deformable
  image registration}.
\newblock \emph{Medical Image Analysis}, 52:\penalty0 128--143, 2019.

\bibitem[Deng et~al.(2009)Deng, Dong, Socher, Li, Li, and
  Fei-Fei]{deng2009imagenet}
Jia Deng, Wei Dong, Richard Socher, Li-Jia Li, Kai Li, and Li~Fei-Fei.
\newblock {Imagenet: A large-scale hierarchical image database}.
\newblock In \emph{Conference on Computer Vision and Pattern Recognition},
  pages 248--255, 2009.

\bibitem[{Di Martino} et~al.(2014){Di Martino}, Yan, Li, and
  Others]{di2014autism}
Adriana {Di Martino}, Chao-Gan Yan, Qingyang Li, and Others.
\newblock {The autism brain imaging data exchange: towards a large-scale
  evaluation of the intrinsic brain architecture in autism}.
\newblock \emph{Molecular psychiatry}, 19\penalty0 (6):\penalty0 659--667,
  2014.

\bibitem[{Faisal Beg} et~al.(2005){Faisal Beg}, Miller, Trouv{\'{e}}trouv, and
  Younes]{faisalBeg2005}
Mirza {Faisal Beg}, Michael~I Miller, Alain Trouv{\'{e}}trouv, and Laurent
  Younes.
\newblock {Computing Large Deformation Metric Mappings via Geodesic Flows of
  Diffeomorphisms}.
\newblock \emph{International Journal of Computer Vision}, 61\penalty0
  (2):\penalty0 139--157, 2005.

\bibitem[{Gal, Yarin and Ghahramani}(2016)]{Gal2016}
Zoubin {Gal, Yarin and Ghahramani}.
\newblock {Dropout as a Bayesian Approximation: Representing Model Uncertainty
  in Deep Learning}.
\newblock In \emph{International Conference on Machine Learning}, pages
  1050----1059, 2016.

\bibitem[Gee et~al.(1993)Gee, Reivich, and Bajcsy]{gee1993elastically}
James~C Gee, Martin Reivich, and Ruzena Bajcsy.
\newblock Elastically deforming a three-dimensional atlas to match anatomical
  brain images.
\newblock \emph{IRCS Technical Reports Series}, \penalty0 (192), 1993.

\bibitem[Haskins et~al.(2019)Haskins, Kruecker, Kruger, Xu, Pinto, Wood, and
  Yan]{Haskins2019a}
Grant Haskins, Jochen Kruecker, Uwe Kruger, Sheng Xu, Peter~A. Pinto, Brad~J.
  Wood, and Pingkun Yan.
\newblock {Learning deep similarity metric for 3D MR–TRUS image
  registration}.
\newblock \emph{International Journal of Computer Assisted Radiology and
  Surgery}, 14\penalty0 (3):\penalty0 417--425, 2019.

\bibitem[Heinrich et~al.(2012)Heinrich, Jenkinson, Bhushan, Matin, Gleeson,
  Brady, and Schnabel]{Heinrich2012}
Mattias~P. Heinrich, Mark Jenkinson, Manav Bhushan, Tahreema Matin, Fergus~V.
  Gleeson, Sir~Michael Brady, and Julia~A. Schnabel.
\newblock {MIND: Modality independent neighbourhood descriptor for multi-modal
  deformable registration}.
\newblock \emph{Medical Image Analysis}, 16\penalty0 (7):\penalty0 1423--1435,
  2012.

\bibitem[Hou et~al.(2017)Hou, Shen, Sun, and Qiu]{hou2017deep}
Xianxu Hou, Linlin Shen, Ke~Sun, and Guoping Qiu.
\newblock {Deep feature consistent variational autoencoder}.
\newblock In \emph{Winter Conference on Applications of Computer Vision}, pages
  1133--1141. IEEE, 2017.

\bibitem[Hu et~al.(2019{\natexlab{a}})Hu, Kang, Huang, Scott, Wiest, and
  Reyes]{Hu2019}
Xiaojun Hu, Miao Kang, Weilin Huang, Matthew~R. Scott, Roland Wiest, and
  Mauricio Reyes.
\newblock {Dual-Stream Pyramid Registration Network}.
\newblock In \emph{International Conference on Medical Image Computing and
  Computer-Assisted Intervention}, pages 382--390. 2019{\natexlab{a}}.

\bibitem[Hu et~al.(2019{\natexlab{b}})Hu, Gibson, Barratt, Emberton, Noble, and
  Vercauteren]{Hu2019a}
Yipeng Hu, Eli Gibson, Dean~C. Barratt, Mark Emberton, J.~Alison Noble, and Tom
  Vercauteren.
\newblock {Conditional Segmentation in Lieu of Image Registration}.
\newblock In \emph{International Conference on Medical Image Computing and
  Computer-Assisted Intervention}, pages 401--409. 2019{\natexlab{b}}.

\bibitem[Ioffe and Szegedy(2015)]{Ioffe2015}
Sergey Ioffe and Christian Szegedy.
\newblock {Batch normalization: Accelerating deep network training by reducing
  internal covariate shift}.
\newblock In \emph{International Conference on Machine Learning}, pages
  448--456. International Machine Learning Society (IMLS), 2015.

\bibitem[Kingma and Ba(2015)]{Kingma2015}
Diederik~P. Kingma and Jimmy~Lei Ba.
\newblock {Adam: A method for stochastic optimization}.
\newblock In \emph{International Conference on Learning Representations}, 2015.

\bibitem[Krebs et~al.(2017)Krebs, Mansi, Delingette, Zhang, Ghesu, Miao, Maier,
  Ayache, Liao, and Kamen]{Krebs2017}
Julian Krebs, Tommaso Mansi, Herv{\'{e}} Delingette, Li~Zhang, Florin~C. Ghesu,
  Shun Miao, Andreas~K. Maier, Nicholas Ayache, Rui Liao, and Ali Kamen.
\newblock {Robust non-rigid registration through agent-based action learning}.
\newblock In \emph{Lecture Notes in Computer Science}, volume 10433, pages
  344--352. Springer Verlag, 2017.

\bibitem[LaMontagne et~al.(2019)LaMontagne, Benzinger, Morris, and
  Others]{lamontagne2019oasis}
Pamela~J LaMontagne, Tammie L~S Benzinger, John~C Morris, and Others.
\newblock {OASIS-3: Longitudinal Neuroimaging, Clinical, and Cognitive Dataset
  for Normal Aging and Alzheimer Disease}.
\newblock \emph{medRxiv}, 2019.

\bibitem[Lee et~al.(2019)Lee, Oktay, Schuh, Schaap, and Glocker]{Lee2019a}
Matthew C.~H. Lee, Ozan Oktay, Andreas Schuh, Michiel Schaap, and Ben Glocker.
\newblock {Image-and-Spatial Transformer Networks for Structure-Guided Image
  Registration}.
\newblock In \emph{International Conference on Medical Image Computing and
  Computer-Assisted Intervention}, pages 337--345. 2019.

\bibitem[Li and Fan(2018)]{Li2018}
Hongming Li and Yong Fan.
\newblock {Non-rigid image registration using self-supervised fully
  convolutional networks without training data}.
\newblock In \emph{Proceedings - International Symposium on Biomedical
  Imaging}, volume 2018-April, pages 1075--1078. IEEE Computer Society, 2018.

\bibitem[Liu et~al.(2019)Liu, Hu, Zhu, and Heng]{Liu2019}
Lihao Liu, Xiaowei Hu, Lei Zhu, and Pheng-Ann Heng.
\newblock {Probabilistic Multilayer Regularization Network for Unsupervised 3D
  Brain Image Registration}.
\newblock In \emph{International Conference on Medical Image Computing and
  Computer-Assisted Intervention}, pages 346--354. 2019.

\bibitem[Ma{\v{s}}ka et~al.(2014)Ma{\v{s}}ka, Ulman, Svoboda, Matula, Matula,
  Ederra, Urbiola, Espa{\~{n}}a, Venkatesan, Balak, Karas, Bolckov{\'{a}},
  {\v{S}}treitov{\'{a}}, Carthel, Coraluppi, Harder, Rohr, Magnusson,
  Jald{\'{e}}n, Blau, Dzyubachyk, Kr{\'{i}}{\v{z}}ek, Hagen, Pastor-Escuredo,
  Jimenez-Carretero, Ledesma-Carbayo, Mu{\~{n}}oz-Barrutia, Meijering, Kozubek,
  and Ortiz-de Solorzano]{Ulman2014}
Martin Ma{\v{s}}ka, Vladim{\'{i}}r Ulman, David Svoboda, Pavel Matula, Petr
  Matula, Cristina Ederra, Ainhoa Urbiola, Tom{\'{a}}s Espa{\~{n}}a,
  Subramanian Venkatesan, Deepak M~W Balak, Pavel Karas, Tereza Bolckov{\'{a}},
  Mark{\'{e}}ta {\v{S}}treitov{\'{a}}, Craig Carthel, Stefano Coraluppi,
  Nathalie Harder, Karl Rohr, Klas E~G Magnusson, Joakim Jald{\'{e}}n, Helen~M
  Blau, Oleh Dzyubachyk, Pavel Kr{\'{i}}{\v{z}}ek, Guy~M Hagen, David
  Pastor-Escuredo, Daniel Jimenez-Carretero, Maria~J Ledesma-Carbayo, Arrate
  Mu{\~{n}}oz-Barrutia, Erik Meijering, Michal Kozubek, and Carlos Ortiz-de
  Solorzano.
\newblock {A benchmark for comparison of cell tracking algorithms}.
\newblock \emph{Bioinformatics}, 30\penalty0 (11):\penalty0 1609--1617, 2014.

\bibitem[Pielawski et~al.(2020)Pielawski, Wetzer, {\"{O}}fverstedt, Lu,
  W{\"{a}}hlby, Lindblad, and Sladoje]{Pielawski2020a}
Nicolas Pielawski, Elisabeth Wetzer, Johan {\"{O}}fverstedt, Jiahao Lu,
  Carolina W{\"{a}}hlby, Joakim Lindblad, and Nata{\v{s}}a Sladoje.
\newblock {CoMIR: Contrastive Multimodal Image Representation for
  Registration}.
\newblock \emph{Advances in neural information processing systems}, 2020.

\bibitem[Quay et~al.(2018)Quay, Emam, Anderson, and Leapman]{Quay2018}
Matthew Quay, Zeyad Emam, Adam Anderson, and Richard Leapman.
\newblock {Designing deep neural networks to automate segmentation for serial
  block-face electron microscopy}.
\newblock In \emph{International Symposium on Biomedical Imaging}, volume
  2018-April, pages 405--408. IEEE Computer Society, 2018.

\bibitem[Ronneberger et~al.(2015)Ronneberger, Fischer, and
  Brox]{Ronneberger2015}
Olaf Ronneberger, Philipp Fischer, and Thomas Brox.
\newblock {U-net: Convolutional networks for biomedical image segmentation}.
\newblock In \emph{International Conference on Medical Image Computing and
  Computer-Assisted Intervention}, volume 9351, pages 234--241. Springer
  Verlag, 2015.

\bibitem[Rueckert et~al.(1999)Rueckert, Sonoda, Hayes, Hill, Leach, and
  Hawkes]{Rueckert1999}
Daniel Rueckert, Luke~I Sonoda, Carmel Hayes, Derek~LG Hill, Martin~O Leach,
  and David~J Hawkes.
\newblock {Nonrigid registration using free-form deformations: Application to
  breast mr images}.
\newblock \emph{IEEE Transactions on Medical Imaging}, 18\penalty0
  (8):\penalty0 712--721, 1999.

\bibitem[Simonovsky et~al.(2016)Simonovsky, Guti{\'{e}}rrez-Becker, Mateus,
  Navab, and Komodakis]{Simonovsky2016}
Martin Simonovsky, Benjam{\'{i}}n Guti{\'{e}}rrez-Becker, Diana Mateus, Nassir
  Navab, and Nikos Komodakis.
\newblock {A deep metric for multimodal registration}.
\newblock In \emph{Lecture Notes in Computer Science}, volume 9902, pages
  10--18. Springer Verlag, 2016.

\bibitem[Thirion(1998)]{thirion1998}
Jean-Philippe Thirion.
\newblock {Image matching as a diffusion process: an analogy with Maxwell's
  demons}.
\newblock Technical Report~3, 1998.

\bibitem[Ulman et~al.(2017)Ulman, Ma{\v{s}}ka, Magnusson, Ronneberger, Haubold,
  Harder, Matula, Matula, Svoboda, Radojevic, Smal, Rohr, Jald{\'{e}}n, Blau,
  Dzyubachyk, Lelieveldt, Xiao, Li, Cho, Dufour, Olivo-Marin, Reyes-Aldasoro,
  Solis-Lemus, Bensch, Brox, Stegmaier, Mikut, Wolf, Hamprecht, Esteves,
  Quelhas, Demirel, Malmstr{\"{o}}m, Jug, Tomancak, Meijering,
  Mu{\~{n}}oz-Barrutia, Kozubek, and Ortiz-De-Solorzano]{Ulman2017}
Vladim{\'{i}}r Ulman, Martin Ma{\v{s}}ka, Klas~E.G. Magnusson, Olaf
  Ronneberger, Carsten Haubold, Nathalie Harder, Pavel Matula, Petr Matula,
  David Svoboda, Miroslav Radojevic, Ihor Smal, Karl Rohr, Joakim Jald{\'{e}}n,
  Helen~M. Blau, Oleh Dzyubachyk, Boudewijn Lelieveldt, Pengdong Xiao, Yuexiang
  Li, Siu~Yeung Cho, Alexandre~C. Dufour, Jean~Christophe Olivo-Marin,
  Constantino~C. Reyes-Aldasoro, Jose~A. Solis-Lemus, Robert Bensch, Thomas
  Brox, Johannes Stegmaier, Ralf Mikut, Steffen Wolf, Fred~A. Hamprecht, Tiago
  Esteves, Pedro Quelhas, {\"{O}}mer Demirel, Lars Malmstr{\"{o}}m, Florian
  Jug, Pavel Tomancak, Erik Meijering, Arrate Mu{\~{n}}oz-Barrutia, Michal
  Kozubek, and Carlos Ortiz-De-Solorzano.
\newblock {An objective comparison of cell-tracking algorithms}.
\newblock \emph{Nature Methods}, 14\penalty0 (12):\penalty0 1141--1152, 2017.

\bibitem[Vercauteren et~al.(2007)Vercauteren, Pennec, Perchant, and
  Ayache]{Vercauteren2007}
Tom Vercauteren, Xavier Pennec, Aymeric Perchant, and Nicholas Ayache.
\newblock {Non-parametric diffeomorphic image registration with the demons
  algorithm}.
\newblock In \emph{Lecture Notes in Computer Science}, volume 4792, pages
  319--326, 2007.

\bibitem[Xu and Niethammer(2019)]{Xu2019}
Zhenlin Xu and Marc Niethammer.
\newblock {DeepAtlas: Joint Semi-supervised Learning of Image Registration and
  Segmentation}.
\newblock In \emph{International Conference on Medical Image Computing and
  Computer-Assisted Intervention}, pages 420--429. 2019.

\bibitem[Yang et~al.(2017)Yang, Kwitt, Styner, and Niethammer]{Yang2017}
Xiao Yang, Roland Kwitt, Martin Styner, and Marc Niethammer.
\newblock {Quicksilver: Fast predictive image registration - A deep learning
  approach}.
\newblock \emph{NeuroImage}, 158:\penalty0 378--396, 2017.

\bibitem[Zhang et~al.(2018)Zhang, Isola, Efros, Shechtman, and Wang]{Zhang2018}
Richard Zhang, Phillip Isola, Alexei~A. Efros, Eli Shechtman, and Oliver Wang.
\newblock {The Unreasonable Effectiveness of Deep Features as a Perceptual
  Metric}.
\newblock \emph{Conference on Computer Vision and Pattern Recognition}, pages
  586--595, 2018.

\end{thebibliography}

\newpage
\appendix
\section{Optimization convergence}
\label{appendix:convergence}

\begin{figure}[H]
\centering
\includegraphics[width=0.8\linewidth]{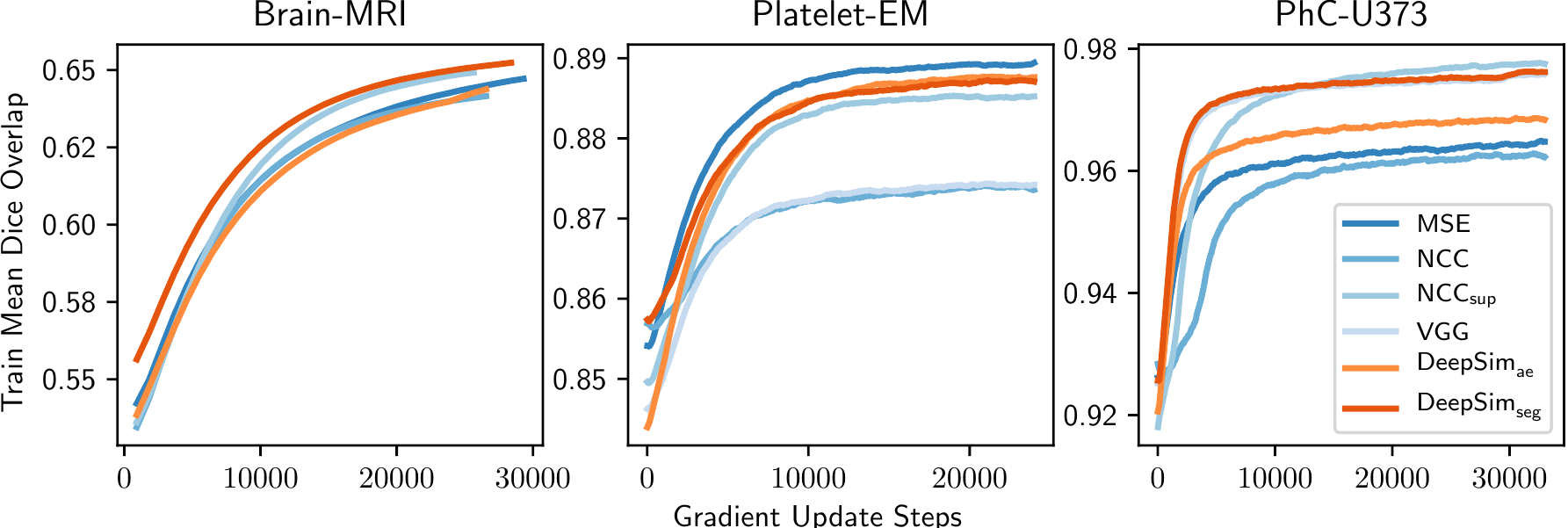}
\caption{Mean dice overlap of the registration models during training. The training duration per model on a single RTX 2080 GPU is approximately seven days for the Brain-MRI dataset and one day for the Platelet, PhC-U373 datasets.}
\label{fig:convergence}
\end{figure}

\section{Regularity of the transformation}
\label{appendix:transformation}

\begin{table}[H]
	\centering
	\begingroup 
	\setlength{\tabcolsep}{3pt} 
\begin{tabular}{@{}lcccccc@{}}
\toprule
Method              & \multicolumn{6}{c}{Dataset}                                                                                                                                                                                                                                                                                                          \\ \cmidrule(r){1-1} \cmidrule(l){2-7}
                    & \multicolumn{2}{c}{Brain-MRI}                                                                              & \multicolumn{2}{c}{Platelet-EM}                                                                            & \multicolumn{2}{c}{PhC-U373}                                                                               \\ \cmidrule(lr){2-3} \cmidrule(lr){4-5} \cmidrule(l){6-7}
                    & \multicolumn{1}{c}{$\sigma^2(J_{\transform})$} & \multicolumn{1}{c}{$|J_{\transform}| < 0 \, (\%)$} & \multicolumn{1}{c}{$\sigma^2(J_{\transform})$} & \multicolumn{1}{c}{$|J_{\transform}| < 0 \, (\%)$} & \multicolumn{1}{c}{$\sigma^2(J_{\transform})$} & \multicolumn{1}{c}{$|J_{\transform}| < 0 \, (\%)$} \\ \addlinespace
$\mse$              & $0.19$                                                & $0.42$                                             & $0.23$                                                & $0.40$                                             & $0.03$                                                & $0.02$                                             \\
$\ncc$              & $0.23$                                                & $0.93$                                             & $0.54$                                                & $4.15$                                             & $0.29$                                                & $0.71$                                             \\
$\ncc_{\text{sup}}$ & $0.11$                                                & $0.28$                                             & $0.54$                                                & $4.03$                                             & $0.28$                                                & $0.57$                                             \\
$\vgg$              & \multicolumn{1}{c}{--}                                & \multicolumn{1}{c}{--}                             & $0.09$                                                & $0.00$                                             & $0.16$                                                & $0.13$                                             \\
$\deepsimae$        & $0.13$                                                & $0.20$                                              & $0.10$                                                & $0.04$                                             & $0.20$                                                & $0.35$                                             \\
$\deepsimseg$       & $0.09$                                                & $0.12$                                             & $0.13$                                                & $0.14$                                             & $0.18$                                                & $0.32$                                             \\ \bottomrule
\end{tabular}
	\endgroup
	\caption{Regularity of the transformation. The determinant of the Jacobian of the transformation $|J_{\transform}|$ is a measure of how the image volume is compressed or stretched by the transformation. We assess transformation smoothness by the variance of the voxel-wise determinant $\sigma^2(J_{\transform})$, a lower variance indicates a more volume-preserving transformation. We assess the regularity of the transformation by measuring the percentage of voxels for which the determinant is $< 0$, which indicates domain folding.}
\end{table}

\begin{figure}[H]
\centering
\includegraphics[width=1\linewidth]{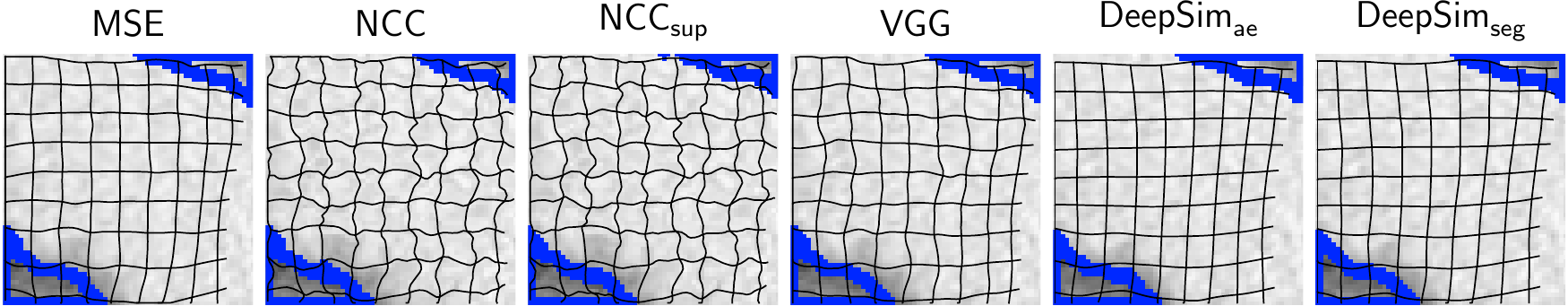}
\caption{Addendum to Figure~\ref{fig:samples_detail}. Detail view of noisy background areas on the Platelet-EM dataset. The cell-boundary is annotated in blue. The transformation is visualized by morphed grid-lines. }
\end{figure}

\end{document}